\documentclass[dvips,???]{imsart}
\usepackage{graphicx}
\usepackage[numbers]{natbib}
\usepackage{amsmath,amsfonts,mathrsfs,bm,graphicx,subfigure,alg}
\usepackage{fullpage}
\usepackage[small,bf]{caption}
\usepackage{algorithmic,multirow}
\usepackage[small,bf]{caption}
\usepackage{graphicx,subfigure}
\usepackage{amsmath,amsfonts,amsthm,mathrsfs,substr,alg}

\RequirePackage[OT1]{fontenc}
\RequirePackage{amsthm,amsmath,natbib}
\RequirePackage[colorlinks,citecolor=blue,urlcolor=blue]{hyperref}
\RequirePackage{hypernat}

\startlocaldefs
\newcommand{\eop}{{\hfill\vbox{\hrule height .2pt
      \hbox{\vrule width.2pt height 6pt
      \kern 4pt
      \vrule width .2pt}
      \hrule height .2pt}} \par\bigskip}
\newtheorem{theorem}{Theorem}

\endlocaldefs

\begin{document}
\begin{frontmatter}

\title{Efficient Latent Variable Graphical Model Selection via Split Bregman Method}
\runtitle{Latent variable graphical model selection}
%\thankstext{T1}{Footnote to the title with the `thankstext' command.}
\begin{aug}
\author{\fnms{Gui-Bo} \snm{Ye},%\thanksref{t1,t2}
\ead[label=e1]{yeg@uci.edu}}
\author{\fnms{Yuanfeng} \snm{Wang},%\thanksref{t1,t2}
\ead[label=e2]{yuanfenw@uci.edu}}
\author{\fnms{Yifei} \snm{Chen},%\thanksref{t1,t2}
\ead[label=e3]{yifeic@uci.edu}}
\and
\author{\fnms{Xiaohui} \snm{Xie}%\thanksref{t3}
\ead[label=e4]{xhx@ics.uci.edu}}
\address{Department of Computer Science, University of California, Irvine\\
Institute for Genomics and Bioinformatics, University of California, Irvine\\
\printead{e4}}

%\thankstext{t1}{Some comment}
%\thankstext{t2}{First supporter of the project}
%\thankstext{t3}{Second supporter of the project}
%\runauthor{G. B. Ye et al.}
\affiliation{University of California Irvine}
\end{aug}

\begin{abstract}
We consider the problem of covariance matrix estimation in the presence of latent variables. Under suitable conditions, it is possible to learn the marginal covariance matrix of the observed variables via a tractable convex program, where the concentration matrix of the observed variables is decomposed into a sparse matrix (representing the graphical structure of the observed variables) and a low rank matrix (representing the marginalization effect of latent variables). We present an efficient first-order method based on split Bregman to solve the convex problem. The algorithm is guaranteed to converge under  mild conditions.   We show that our algorithm is significantly faster than the state-of-the-art algorithm on both artificial and real-world data. Applying the algorithm to a gene expression data involving thousands of genes, we show that most of the correlation between observed variables can be explained by only a few dozen latent factors.
\end{abstract}

\begin{keyword}[class=AMS]
\kwd[Applied statistics ]{97K80}
\kwd[; Learning and adaptive systems ]{68T05}
\kwd[; Convex programming ]{90C25}
\end{keyword}

\begin{keyword}
\kwd{Gaussian graphical models}
\kwd{Variable selection }
\kwd{Alternative direction method of multipliers}
\end{keyword}
%\tableofcontents
\end{frontmatter}

\section{Introduction}
%A graphical model is a statistical model defined with respect to a graph $(V,\mathcal{E})$ in which the nodes index a collection of random variables $\{X_v\}_{v\in V}$, and the edges represent the conditional independence relations among the variables. The absence of an edge between nodes $i,j\in V$  implies that the variables $X_i,X_j$ are independent conditioned on all the other variables. An important class of undirected graphical model is Gaussian grphical model (GGM) in which variables in the model jointly follows a multi-variate gaussian distribution.

Estimating covariance matrices in the high-dimensional setting arises in many applications and has drawn considerable interest recently. Because the sample covariance matrix is typically poorly behaved in the high-dimensional regime, regularizing the sample covariance based on some assumptions of the underlying true covariance is often essential to gain robustness and stability of the estimation.

One form of regularization that has gained popularity recently is to require the the underlying inverse covariance matrix to be sparse \cite{Dempster:Biometrics:1972,PWZZ:JASA:2009,BEd:JMLR:2008,FHT:Biostatistics:2008}. If the data follow a multivariate Gaussian distribution with covariance matrix $\Sigma$, the entries of the inverse covariance matrix $K=\Sigma^{-1}$ (also known as concentration matrix or precision
matrix) encode the information of conditional dependencies between variables: $K_{ij}=0$ if the variables $i$ and $j$ are conditionally independent given all others. Therefore, the sparsity regularization is equivalent to the assumption that most of the variable pairs in the high-dimensional setting are conditionally independent.

To make the estimation problem computational tractable, one often adopts a convex relaxation of the sparsity constraint and uses the $\ell_1$ norm to promote the sparsity of the concentration matrix \cite{BEd:JMLR:2008,FHT:Biostatistics:2008,MB:AS:2006,YL:Biometrika:2007}. Denote $\Sigma^n$ the empirical covariance.  Under the maximum likelihood framework, the covariance matrix estimation problem is then formulated as solving the following optimization problem:
\begin{align}\label{min l1 penalized likelihood}
&\min -\log\det K+\hbox{tr}(\Sigma^n K)+\lambda\|K\|_1\nonumber\\
 &{\rm s.t.} \quad K\succeq 0,
\end{align}
where tr denotes the trace, $\lambda$ is a sparsity regularization parameter, and $K\succeq 0$ denotes that $K$ is positive semidefinite. Due to the $\ell_1$ penalty term and the explicit positive definite constraint on $K$, the method leads to a sparse estimation of the concentration matrix that is guaranteed to be positive definite. The problem is convex and many algorithms have been proposed to solve the problem efficiently in high dimension \cite{FHT:Biostatistics:2008,Lu:SIAM:2008,SMG:NIPS:2010,Yuan:preprint:2009}.

However, in many real applications only a subset of the variables are directly observed, and no additional information is provided on both the number of the latent variables and their relationship with the observed ones.
For instance, in the area of functional genomics it is often the case that only mRNAs of the genes can be directly measured, but not the proteins, which are correlated but have no direct correspondence to the mRNAs because of the prominent role of the postranscriptional regulation. Another example is the movie recommender system where the preference of a movie can be strongly influenced by latent factors such as advertisements, social environment, etc. In these and other cases, the observed variables can be densely correlated because of the marginalization over the unobserved hidden variables. Therefore, the sparsity regularization alone may fail to achieve the desire results.

We consider the setting in which the hidden ($X_H$) and the observed variables ($X_O$) are jointly Gaussian with covariance matrix $\Sigma_{(O H)}$. The marginal statistics of the observed variable $X_O$ are given by the marginal covariance matrix $\Sigma_{O}$, which is simply a submatrix of the full covariance matrix $\Sigma_{(O H)}$. Let the concentration matrix $K_{(O H)}=\Sigma^{-1}_{(O H)}$. The marginal concentration matrix $\Sigma_O^{-1}$ corresponding to the observed variables $X_O$ is given by the Schur complement  \cite{CPW:CCC:2010}:
\begin{equation}\label{marginal precision_matrix}
\hat{K}_O=\Sigma_{O}^{-1} = K_O - K_{O,H} K_H^{-1} K_{H,O},
\end{equation}
where $K_O, K_{O,H}$, and $K_H$ are the corresponding submatrices of the full concentration matrix. Based on the Schur complement, it is clear that the marginal concentration matrix of the observed variables can be decomposed into two components: one is $K_O$, which specifies the conditional dependencies of the observed variables given both the observed and latent variables, and the other is $K_{O,H} K_H^{-1} K_{H,O}$, which represents the effect of marginalization over the hidden variables.  One can now impose assumptions to the two underlying components separately.

By assuming that the $K_O$ matrix is sparse and the number of latent variables is small, the maximum likelihood estimation of the covariance matrix of the observed variables at the presence of latent variables can then be formulated as
\begin{align}\label{Problem formulation}
&\min_{S,L} -\log\det(S-L)+\operatorname{tr}(\Sigma_{O}^{n}(S-L))+ \lambda_1 \|S\|_1 + \lambda_2 \operatorname{tr}(L) \nonumber\\
&{\rm s.t.}\quad S-L\succeq 0,\quad L\succeq 0.
\end{align}
where we decompose $\Sigma_O^{-1} = S-L$ with $S$ denoting $K_O$ and $L$ denoting $K_{O,H} K_H^{-1} K_{H,O}$. Because the number of the hidden variables is small, $L$ is of low rank, whose convex relaxation is the trace norm. There are two regularization parameters in this model: $\lambda_1$ regularizes the sparsity of $S$, and
$\lambda_2$ regularizes the rank of $L$. Under certain regularity conditions, Chandrasekaran et al. showed that this model can consistently estimate the underlying model structure in the high-dimensional regime in which the number of observed/hidden variables grow with the number of samples of the observed variables \cite{CPW:CCC:2010}.

The objective function in \eqref{Problem formulation} is strictly convex, so a global optimal solution is guaranteed to exist and be unique. Finding the optimal solution in the high-dimension setting is computationally challenging due to  the $\log\det$ term appeared in the likelihood function, the trace norm, the nondifferentiability of  the $\ell_1$ penalty, and the positive semidefinite constraints. For large-scale problems, the state-of-the-art algorithm for solving \eqref{Problem formulation} is to use the special purpose solver LogdetPPA \cite{WST:SIAM:2010} developed for log-determinant semidefinite programs. However, the solver LogdetPPA is designed to solve smooth problems. In order to use LogdetPPA, one has to reformulate \eqref{Problem formulation} to a smooth problem. As a result, no optimal sparse matrix $S$ can be generated and additional heuristic steps involving thresholding have to be applied in order to produce sparsity. In addition, LogdetPPA is not especially designed for \eqref{Problem formulation}. We believe much more efficient algorithms can be generated by considering the unique structures of the model specifically.

The main contribution of this paper contains two aspects. First, we present a new algorithm for solving \eqref{Problem formulation} and show that the algorithm is significantly faster than the state-of-the-art method, especially for large-scale problems.  The algorithm is derived by reformulating the problem and adapting the split Bregman method \cite{SMG:NIPS:2010,Yuan:preprint:2009}.  We derive closed form solutions for each subproblem involved in the split Bregman iterations. Second, we apply the method to analyze a large-scale gene expression data, and find that the model with latent variables explain the data much better than the one without assuming latent variables. In addition, we find that most of the correlations between genes can be explained by only a few latent factors, which provides a new aspect for analyzing this type of data.

The rest of the paper is organized as follows. In Section \ref{sec split Bregman method for LVGG}, we derive a split Bregman method, called SBLVGG, to solve the latent variable graphical model selection problem \eqref{Problem formulation}. The convergence property of the algorithm is also given. SBLVGG consists of four update steps and each update has explicit formulas to calculate. In Section \ref{sec experiment}, we illustrate the utility of our algorithm and compare its performance to LogdetPPA using both simulated data and gene expression data.

\section{Split Bregman method for latent variable graphical model selection}\label{sec split Bregman method for LVGG}
The split Bregman method was originally proposed by Osher and coauthors to solve total variation based image restoration problems \cite{GO:SIAM:2009}. It was later found to be either equivalent or closely related to a number of other existing optimization algorithms, including Douglas-Rachford splitting \cite{WuTai:SIJISC:2010}, the alternating direction method of multipliers (ADMM) \cite{GabayMercier:CMA:1976,GlowinskiMarrocco:RFAIRO:1975,GO:SIAM:2009} and the method of multipliers \cite{Rockafellar:MP:1973}. Because of its fast convergence and the easiness of implementation, it is increasingly becoming a method of choice for solving large-scale sparsity recovery problems \cite{CaiOsherShen:MMS:2009,CLMW:Arxive:2009}. Recently, it is also used to solve \eqref{min l1 penalized likelihood} and find it is very successful \cite{SMG:NIPS:2010,Yuan:preprint:2009}.

In this section, we first  reformulate the problem by introducing an auxiliary variable and then proceed to derive a split Bregman method to solve the reformulated problem. Here we would like to emphasize that, although split Bregman method has been introduced to solve graphical model problems \cite{SMG:NIPS:2010,Yuan:preprint:2009}, we have our own contributions. Firstly, it is our first time to use split Bregman method to solve \eqref{Problem formulation} and we introduce an auxiliary variable for a data fitting term instead of penalty term which has been adopted in \cite{SMG:NIPS:2010,Yuan:preprint:2009}. Secondly,
Secondly, the update three hasn't been appeared in \cite{SMG:NIPS:2010,Yuan:preprint:2009} and we provide an explicit formula for it as well. Thirdly, instead of using eig (or schur) decomposition as done in previous work \cite{SMG:NIPS:2010,Yuan:preprint:2009}, we use the LAPACK routine dsyevd.f (based on a divide-and-conquer strategy) to compute the full eigenvalue decomposition of a symmetric matrix which is essential for updating the first and third subproblems.
\subsection{Derivation of the split bregman method for latent variable graphical model selection}\label{subsec framework}
The log-likelihood term and the regularization terms in \eqref{Problem formulation} are coupled, which makes the optimization problem difficult to solve. However, the three terms can be decoupled if we introduce an auxiliary variable to transfer the coupling from the objective function to the constraints.  More specially, the problem \eqref{Problem formulation} is equivalent to the following  problem
\begin{eqnarray}\label{reformulation}
(\hat{A}, \hat{S}, \hat{L} )&=& \arg\min_{A, S,L}-\log\det A+tr(\Sigma_{O}^{n}A) + \lambda_1 \|S\|_1 + \lambda_2 tr(L) \\
  s.t. && A = S-L  \nonumber\\
       && A\succ 0, L\succeq0 \nonumber.
\end{eqnarray}
The introduction of the new variable of $A$ is a key step of our algorithm, which makes the problem amenable to a split Bregman procedure to be detailed below.
Although the split Bregman method originated from Bregman iterations, it has been demonstrated to be equivalent to the alternating direction method of multipliers (ADMM) \cite{GabayMercier:CMA:1976,GlowinskiMarrocco:RFAIRO:1975,Setzer:SSVMCV:2009}. For simplicity of presentation, next we derive the split Bregman method using the augmented Lagrangian method \cite{Hestenes:JOTA:1969,Rockafellar:MP:1973}.

We first define an augmented Lagrangian function of \eqref{reformulation}
\begin{eqnarray}\label{eq Aug Lag}
\mathcal{L}(A,S,L,U)&:=& -\log\det A+tr(\Sigma_{O}^{n}A) + \lambda_1 \|S\|_1 + \lambda_2 tr(L)\nonumber\\&& + tr(U(A-S+L)) + \frac{\mu}{2} \|A-S+L\|^2_F,
\end{eqnarray}
where $U$ is a dual variable matrix corresponding to the equality constraint $A = S-L$, and $\mu>0$ is a parameter. Compared with the standard Lagrangian function, the augmented Lagrangian function has an extra term $\frac{\mu}{2} \|A-S+L\|^2_F$, which penalizes the violation of the linear constraint $A = S-L$.

With the definition of the augmented Lagrangian function \eqref{eq Aug Lag}, the primal problem \eqref{reformulation} is equivalent to
\begin{equation}\label{min primal}
  \min_{A\succ 0,L\succeq 0,S}\max_{U}\mathcal{L}(A,S,L,U).
\end{equation}
Exchanging the order of $\min$ and $\max$ in \eqref{min primal} leads to the formulation of the dual problem
\begin{equation}\label{max dual}
  \max_{U}E(U)\quad\hbox{with}\quad E(U)=\min_{A\succ 0,L\succeq 0,S}\mathcal{L}(A,S,L,U).
\end{equation}
Note that the gradient $\nabla E(U)$ can be calculated by the following \cite{Bertsekas:book:1982}
\begin{equation}\label{eq gradient}
  \nabla E(U)= A(U)-S(U)+L(U),
\end{equation}
where $(A(U), S(U),L(U))=\arg\min_{A\succ 0,L\succeq 0,S}\mathcal{L}(A,S,L,U)$.

Applying gradient ascent on the dual problem \eqref{max dual} and using equation \eqref{eq gradient}, we obtain the method of multipliers \cite{Rockafellar:MP:1973} to solve \eqref{reformulation}
\begin{equation}\label{eq method of multiplier}
 \begin{cases}
 (A^{k+1}, S^{k+1}, L^{k+1}) = \arg\min_{A\succ 0,L\succeq 0,S} \mathcal{L}(A,S,L,U^k), \\
 U^{k+1}=U^k+\mu(A^{k+1}-S^{k+1}+L^{k+1}).
 \end{cases}
\end{equation}
Here we have used $\mu$ as the step size of the gradient ascent. It is easy to see that the efficiency of the iterative algorithm \eqref{eq method of multiplier} largely hinges on whether the first equation of \eqref{eq method of multiplier} can be solved efficiently. Note that the  augmented Lagrangian function $\mathcal{L}(A,S,L,U^k)$ still contains $A,S,L$ and can not easily be solved directly. But we can solve the first equation of \eqref{eq method of multiplier} through an iterative algorithm that alternates between the minimization of $A,S$ and $L$. The method of multipliers requires that the alternative minimization of $A,S$ and $L$ are run multiple times until convergence to get the solution  $(A^{k+1}, S^{k+1}, L^{k+1})$.  However, because  the first equation of \eqref{eq method of multiplier} represents only one step of the overall iteration, it is actually not necessary to be solved completely.  In fact, the split Bregman method (or the alternating direction method of multipliers \cite{GabayMercier:CMA:1976}) uses only one alternative iteration to get a very rough solution of  \eqref{eq method of multiplier}, which leads to the following iterative algorithm for solving \eqref{reformulation} after some reformulations,
\begin{equation} \label{eq SBLVGG}
\begin{cases}
A^{k+1}=\arg\min_{A\succ 0}  -\log\det(A)+ tr(A \Sigma_{O}^{n}) + \frac{\mu}{2} \big\|A-S^k+L^k+\frac{U^k}{\mu}\big\|^2_F, \\
S^{k+1}=\arg\min_{S}  \lambda_1 \|S\|_1 + \frac{\mu}{2} \big\|A^{k+1}-S+L^k+\frac{U^k}{\mu}\big\|^2_F,\\
L^{k+1}=\arg\min_{L\succeq0}  \lambda_2 tr(L) +\frac{\mu}{2} \big\|A^{k+1}-S^{k+1}+L+\frac{U^k}{\mu}\big\|^2_F,\\
U^{k+1}=U^k+\mu(A^{k+1}-S^{k+1}+L^{k+1}).
\end{cases}
\end{equation}
\subsubsection{Convergence}
The convergence of the iteration \eqref{eq SBLVGG}  can be derived from the convergence theory of the alternating direction method of multipliers or the convergence theory of the split Bregman method \cite{GabayMercier:CMA:1976,EcksteinDouglas:MP:1992,CaiOsherShen:MMS:2009}.
\begin{theorem}\label{theorem convergence}
Let $(S^k,L^k)$ be generated by \eqref{eq SBLVGG}, and $(\hat{S},\hat{L})$ be the unique minimizer of \eqref{reformulation}. Then,
\begin{equation*}
    \lim_{k\rightarrow\infty}\|S^k-\hat{S}\|=0\quad\hbox{and}\quad \lim_{k\rightarrow\infty}\|L^k-\hat{L}\|=0.
  \end{equation*}
\end{theorem}
From Theorem \ref{theorem convergence}, the condition for the convergence of the iteration \eqref{eq SBLVGG} is quite mild and even irrelevant to the choice of the parameter $\mu$ in the iteration \eqref{eq SBLVGG}.
\subsubsection{Explicit formulas to update $A,S$ and $L$}
We first focus on the computation of the first equation of \eqref{eq SBLVGG}. Taking the derivative of the objective function and setting it to be zero, we get
\begin{equation}\label{eq A derivative}
  -A^{-1}+\Sigma_{O}^{n}+U^k+\mu(A-S^k+L^k)=0
\end{equation}
It is a quadratic equation where the unknown is a matrix. The complexity for solving this equation is at least $O(p^3)$ because of the inversion involved in \eqref{eq A derivative}. Note that $\|S\|_1=\|S^T\|_1$ and $L=L^T$, if $U^k$ is symmetric, so is $\Sigma_{O}^{n}+U^k-\mu(S^k-L^k)$.
It is easy to check that the explicit form for the solution of \eqref{eq A derivative} under constraint $A \succ 0$, i.e., $A^{k+1}$, is
\begin{equation}\label{eq update A}
A^{k+1} = \frac{K^k+\sqrt{(K^k)^2+4\mu I}}{2\mu},
\end{equation}
where $K^k = \mu(S^k-L^k)-\Sigma_{O}^{n}-U^k$ and $\sqrt{C}$ denotes the square root of a symmetric positive definite matrix $C$. Recall that the square root of a symmetric positive definite matrix $C$ is defined to be the matrix whose eigenvectors are the same as those of $C$ and eigenvalues are the square root of those of $C$. Therefore, to get the update of $A^{k+1}$, can first compute the eigenvalues and eigenvectors of $K^k$, and then get the eigenvalues of $A^{k+1}$ according to \eqref{eq update A} by replacing the matrices by the corresponding eigenvalues. We adopt the LAPACK routine dsyevd.f (based on a divide-and-conquer strategy) to compute the full eigenvalue decomposition of $(K^k)^2+4\mu I$. It is about $10$ times faster than eig (or schur) routine when $n$ is larger than $500$.

For the second equation of \eqref{eq SBLVGG}, we have made the data fitting term $\frac{\mu}{2} \big\|A^{k+1}-S+L^k+\frac{U^k}{\mu}\big\|^2_F$ separable with respect to the entries of $A$. Thus, it is very easy to get the solution and the computational complexity would be $O(p^2)$ for $\|S\|_1$ is also separable.  Let $\mathcal{T}_\lambda$ be a soft thresholding operator defined on matrix space and satisfying
$$\mathcal{T}_\lambda(\Omega)=(t_\lambda(\omega_{ij}))_{i,j=1}^p,$$ where $t_\lambda(\omega_{ij})=\hbox{sgn}(\omega_{ij})\max\{0,|\omega_{ij}|-\lambda\}$.
Then the update of $S$ is
$$S^{k+1}=\mathcal{T}_{\frac{\lambda_1}{\mu}}(A^{k+1}+L^k+\mu^{-1} U^k).$$

For the update of $L$, it can use the following Theorem.
\begin{theorem}\label{theorem projected SVT}
Given a symmetric matrix $X$ and $\eta>0$. Denote $$\mathcal{S}_\eta(X)=\arg\min_{Y\succeq 0} \eta tr(Y) + \frac{1}{2} \|Y-X\|^2_F.$$ Then
$\mathcal{S}_\eta(X)=V diag((\lambda_i-\eta)_+) V^T$, where $\lambda_i (i\in{1,...,n})$ are the eigenvalues of $X$ with $V$ being the corresponding eigenvector matrix and $(\lambda_i-\eta)_+=\max(0,\lambda_i-\eta)$.
\end{theorem}
\proof Note that $tr(Y)=\langle I, Y \rangle,$ where $I$ is the identity matrix. Thus, $\arg\min_{Y\succeq 0} \eta tr(Y) + \frac{1}{2} \|Y-X\|^2_F=\arg\min_{Y\succeq 0} \langle Y-X+\eta I, Y-X+\eta I\rangle.$
Compute eigenvalue decomposition on matrix $X$ and get $X = V \Lambda V^T$, where $VV^T=V^TV=I$ and $\Lambda$ is the diagonal matrix. Then
\begin{equation*}
  \langle Y-X+\eta I, Y-X+\eta I \rangle = \langle V^T Y V - (\Lambda -\eta I), V^T Y V - (\Lambda -\eta I) \rangle.
\end{equation*}
 Together with the fact that $\mathcal{S}_\eta(X)\succeq 0$,  $\mathcal{S}_\eta(X)$ should satisfy $(V^T \mathcal{S}_\eta(X) V)_{ij} =  \max(0,\lambda_i-\eta)$  for $i=j$ and $0$ otherwise.
Therefore, $\mathcal{S}_\eta(X)=V diag((\lambda_i-\eta)_+) V^T$.
\eop
Using the operator $\mathcal{S}_\eta$ defined in Theorem \ref{theorem projected SVT}, it is easy to see that
\begin{equation}
L^{k+1}=\mathcal{S}_{\frac{\lambda_2}{\mu}}(S^{k+1}-A^{k+1}-\mu^{-1}U^k).
\end{equation}
Here we also use the LAPACK routine dsyevd.f (based on a divide-and-conquer strategy) to compute the full eigenvalue decomposition of $S^{k+1}-A^{k+1}-\mu^{-1}U^k$.
Summarizing all together, we get SBLVGG to solve the latent variable Gaussian Graphical Model \eqref{Problem formulation} as shown in Algorithm 1.
\begin{algorithm}[htb]\label{alg-SBLVGG}
\caption{Split Bregman method for solving Latent Variable Gaussian Graphical Model (SBLVGG)} 
\begin{algorithmic}
	\STATE Initialize $S^0,L^0, U^0$.
	\REPEAT
        \STATE 1) $A^{k+1} = \frac{K^k + \sqrt{(K^k)^2+4\mu I}}{2\mu}$, where $K^k = \mu(S^k-L^k) - \Sigma - U^k$
		\STATE 2) $S^{k+1} = \mathcal{T}_{\frac{\lambda_1}{\mu}} (A^{k+1} + L^k + \mu^{-1}U^k)$
		\STATE 3) $L^{k+1}=\mathcal{S}_{\frac{\lambda_2}{\mu}}(S^{k+1}-A^{k+1}-\mu^{-1}U^k)$
		\STATE 4) $U^{k+1} = U^k+\mu (A^{k+1}-S^{k+1}+L^{k+1})$
	\UNTIL{\STATE Convergence}
\end{algorithmic}
\end{algorithm}

\section{Numerical experiments}\label{sec experiment}
Next we illustrate the efficiency of the split Bregman method (SBLVGG) for solving \eqref{Problem formulation} using time trials on artificial data as well as gene expression data. All the algorithms were implemented in Matlab and run on a 64-bit linux desktop with Intel i3 - 3.2GHz QuadCore CPU and 8GB memory. To evaluate the performance of SBLVGG, we compare it with logdetPPA \cite{WST:SIAM:2010} which is state-of-art solver for \eqref{Problem formulation} in large-scale case. LogdetPPA was originally developed for log-determinant semidefinite programs with smooth penalties. In order to solve \eqref{Problem formulation} using LogdetPPA , we need to reformulate \eqref{Problem formulation} as a smooth problem as done in \cite{CPW:CCC:2010}, which results in the derived sparse matrix $\hat{S}$  not strictly sparse with many entries close to but not exactly $0$. We also demonstrate that latent variable Gaussian graphical selection model \eqref{Problem formulation} is better than sparse Gaussian graphical model \eqref{min l1 penalized likelihood} in terms of generalization ability using gene expression data.

Note that the convergence of Algorithm \ref{algorithm SBLVGG} is guaranteed no matter what values of $\mu$ is used as shown in Theorem \ref{theorem convergence}. The speed of the algorithm can, however, be influenced by the choices of $\mu$ as it would affect the number of iterations involved. In our implementation, we choose $\mu$ in  $[0.005,0.01]$ for artificial data and $[0.001,0.005]$ for gene expression data.
\subsection{Artificial data}
Let $p = p_o+p_h$ with $p$ being the total number of variables in the graph, $p_o$  the number of observed variables and $p_h$  the number of hidden variables. The synthetic data are generated in a similar way as the one in Section 6.1 of \cite{WST:SIAM:2010}. First, we generate an $p\times p$ random sparse matrix $W$ with non-zero entries drawn from normal distribution $\mathcal{N}(0,1)$. Then set
\begin{eqnarray*}
&&C = W'*W; \quad C(1:p_o, p_o+1:p) = C(1:p_o, p_o+1:p)+0.5*randn(p_o,p_h);\\
&&C = (C+C')/2;\quad d=diag(C);\quad C=max(min(C-diag(d),1),-1);\\
&&K  = B+max(-1.2*min(eig(B)), 0.001)*eye(p);\quad K_O = K(1:p_o,1:p_o)\\
&&K_{OH} = K(1:p_o,p_o+1:p);\quad K_{HO} = K(p_o+1:p,1:p_o);\\
&&K_H = K(p_o+1:p,p_o+1:p);\tilde{K}_O = K_O - K_{OH}K_H^{-1}K_{HO}.
\end{eqnarray*}
\begin{figure}[h]
\centering
\subfigure[]{\includegraphics[width=.45\textwidth]{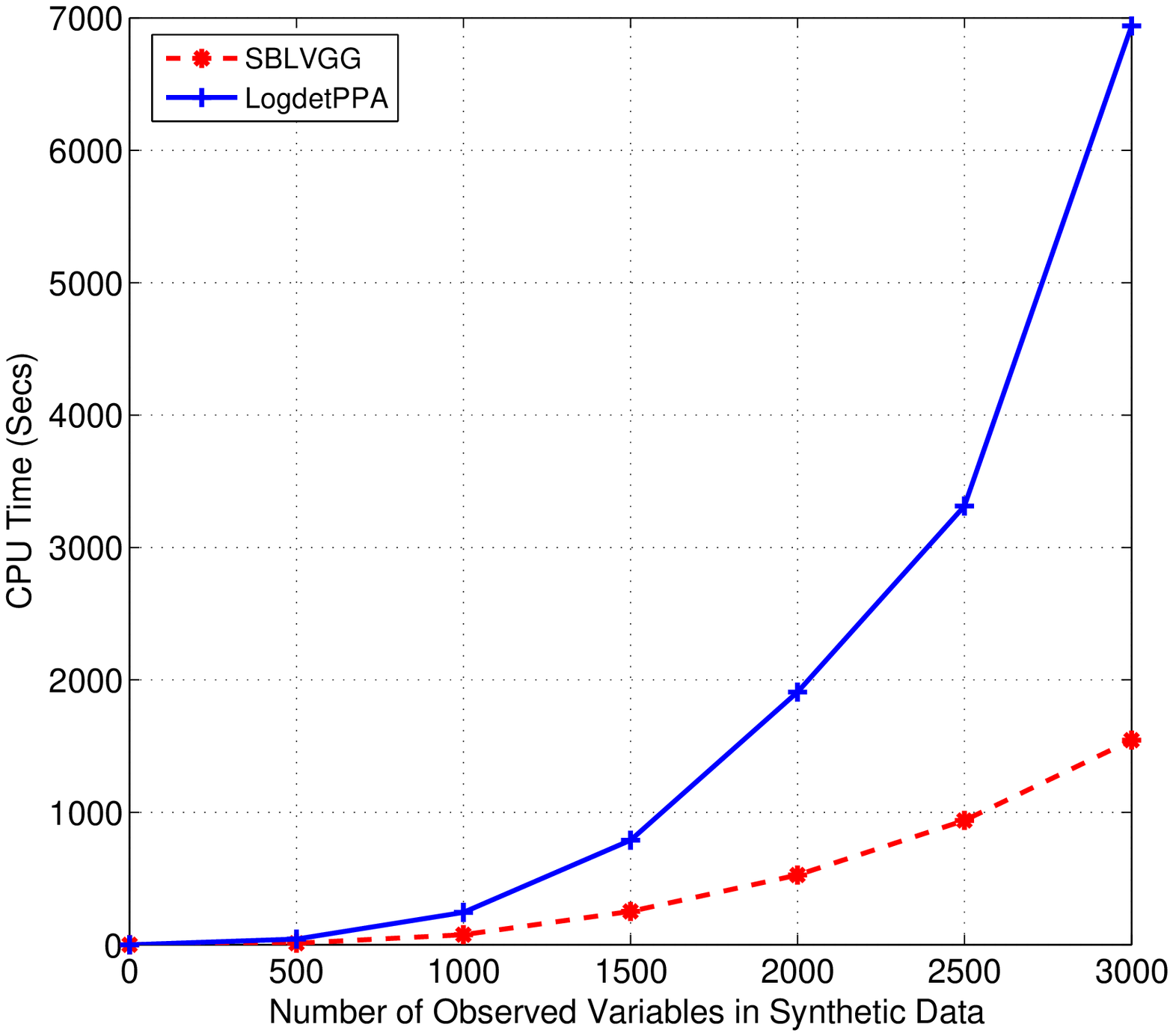}}
\subfigure[]{\includegraphics[width=.45\textwidth]{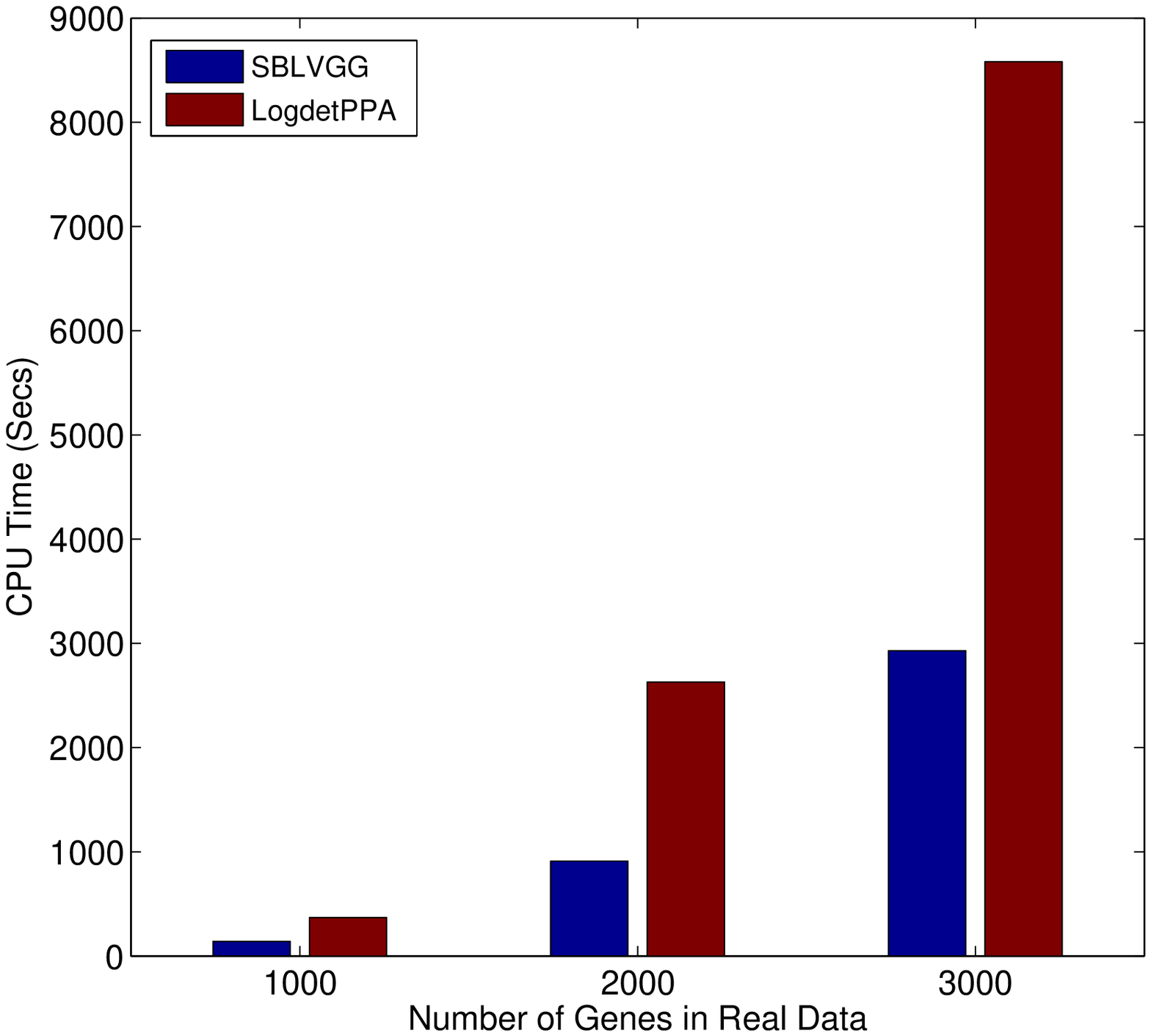}}
\caption{(a) Comparison of CPU time curve w.r.t. number of variables $p$ for artificial data; (b) Comparison of CPU time curve w.r.t. number of variables $p$ for gene expression data \label{Fig:CPU:Time}}
\end{figure}

%\begin{table}
%\caption{Parameter selection of $\lambda_1,\lambda_2$ at different problem scales for simulation data}
%\label{Model:Selection}
%\begin{center}
%\begin{tabular}{ l|llll}
%\multirow{2}{*}{Problem Scale} & \multicolumn{4}{c}{$(\lambda_1, \lambda_2)$ Pairs}  \\ \cline{2-5}
%	  		& Pair 1 		& Pair 2 		& Pair 3 		& Pair 4 \\ \hline
%(500, 10)  	& 0.005, 0.096 	& 0.004, 0.08	& 0.004, 0.096 	& 0.0032, 0.08 	\\
%(1000, 10)  	& 0.0028, 0.10	& 0.0028, 0.12	& 0.0034, 0.10	& 0.0022, 0.10 	\\
%(1500, 10)  	& 0.003, 0.17 	& 0.0035, 0.17	& 0.0035, 0.18 	& 0.004, 0.18 \\
%(2000, 10)  	& 0.002, 0.10 	& 0.002, 0.12	& 0.0024, 0.10 	& 0.0016, 0.10 \\
%(2500, 10)  	& 0.0021, 0.17	& 0.0023, 0.17	& 0.0025, 0.17 	& 0.0025, 0.18\\
%(3000, 10)  	& 0.0025, 0.21 	& 0.0025, 0.22 	& 0.0027, 0.21 	& 0.0027, 0.22\\
%\end{tabular}
%\end{center}
%\end{table}
\begin{table}
\caption{Numerical comparison at $p_o=3000, p_h=10$ for artificial data}
\label{Tab:Acc:3000}
\begin{center}
\begin{tabular}{ c|c|c|c|c}
($\lambda_1, \lambda_2$)& Method		& Obj. Value 				& Rank	& Sparse Ratio	\\ \hline
\multirow{2}{*}{(0.0025, 0.21)} & SBLVGG 	& -5642.6678		& 8		& 5.56\%		\\
			           & lodgetPPA 	& -5642.6680		& 8		& 99.97\%		 \\ \hline
\multirow{2}{*}{(0.0025, 0.22)} & SBLVGG 	& -5642.4894		& 3		& 5.58\%		\\
			           & lodgetPPA 	& -5642.4895		& 3		& 99.97\%		 \\ \hline
\multirow{2}{*}{(0.0027, 0.21)} & SBLVGG	& -5619.2744		& 16		& 4.14\%		 \\
			           & lodgetPPA 	& -5619.2746		& 16		& 99.97\%		 \\ \hline
\multirow{2}{*}{(0.0027, 0.22)} & SBLVGG 	& -5619.0194		& 6		& 4.17\%		\\
			           & lodgetPPA	& -5619.0196		& 6		& 99.97\%		 \\ \hline			
\end{tabular}
\end{center}
\end{table}

Note that $\tilde{K}_O$ is marginal precision matrix of observed variables. We generate $n$ Gaussian random samples from $\tilde{K}_O$, and calculates its sample covariance matrix $\Sigma_O^n$. In our numerical experiments, we set sparse ratio of $K_O$ around $5\%$, and $p_h=10$. The stopping criteria for SBLVGG is specified as follows. Let $\Phi(A,L) = -\log\det A + tr(A\Sigma) + \lambda_1 \|A+L\|_1 + \lambda_2 tr(L)$. We stop our algorithm if
$|\Phi(A^{k+1},L^{k+1}) - \Phi(A^k,L^k)|/\max \left(1, |\Phi(A^{k+1},L^{k+1})| \right) < \epsilon$ and $\|A-S+L\|_F < \epsilon$ with $\epsilon=1e-4$.

Figure \ref{Fig:CPU:Time}(a) shows CPU time curve of SBLVGG and LogdetPPA with respect to the number of variable $p$ for the artificial data. For each fixed $p$, the CPU time is averaged over 4 runs with four different $(\lambda_1, \lambda_2)$ pairs. We can see SBLVGG consistently outperform LogdetPPA. For dimension of 2500 or less, it is 3.5 times faster on average. For dimension 3000, it is 4.5 times faster. This also shows SBLVGG scales better to problem size than LogdetPPA. In terms of accuracy, Table \ref{Tab:Acc:3000} summarize performance of two algorithms at $p_o=3000$, $p_h=10$ in three aspects: objective value, rank of $L$, sparsity of $S$ (ratio of non-zero off-diagonal elements). We find in terms of objective value and rank, both algorithms generate almost identical results. However, SBLVGG outperform LogdetPPA due to its  soft- thresholding operator in Algorithm \ref{algorithm SBLVGG} for $S$, while LogdetPPA misses this kind of operator and result in many nonzero but close to zero entries  due to numerical error. We would like to emphasize that the results in lower dimensions are very similar to $p_o=3000$, $p_h=10$. We omit the details here due space limitation.
\subsection{Gene expression data}
The gene expression data \cite{Hughes:Cell:2000} contains measurements of mRNA expression level of the $6316$ genes of {\it S. cerevisiae} (yeast) under $300$ different experimental conditions. First we centralize the data and choose three subset of the data, $1000$, $2000$ and $3000$ genes with highest variances. Figure \ref{Fig:CPU:Time}(b) shows CPU time of SBLVGG and LogdetPPA with different $p$.  We can see that SBLVGG consistently perform better than LogdetPPA: in 1000
\begin{table}[!h]
\caption{Numerical comparison at 3000 dimensional subset of gene expression data}
\label{Tab:Acc:3000:Gene:Data}
\begin{center}
\begin{tabular}{ c|c|c|c|c}
($\lambda_1,\lambda_2$)	& Algorithm	& Obj. Value 			& Rank	&\# Non-0 Entries	\\ \hline
\multirow{2}{*}{(0.01,0.05)} 	& SBLVGG 	& -9793.3451				& 88		 & 34		 \\
			           		& LodgetPPA 	& -9793.3452				& 88		 & 8997000	\\ \hline
\multirow{2}{*}{(0.01,0.1)} 	& SBLVGG 	& -9607.8482				& 60		 & 134		 \\
			           		& LodgetPPA 	& -9607.8483				& 60		 & 8997000		 \\ \hline
\multirow{2}{*}{(0.02,0.05)} 	& SBLVGG 	& -8096.2115				& 79		 & 0		 \\
			           		& LodgetPPA 	& -8096.2115				& 79		 & 8996998		 \\ \hline
\multirow{2}{*}{(0.02,0.1)} 	& SBLVGG 	& -8000.9047				& 56		 & 0		 \\
			           		& LodgetPPA	& -8000.9045				& 56		 & 8997000	\\ \hline			
\end{tabular}
\end{center}
\end{table}
dimension case, SBLVGG is $2.5$ times faster,  while in 2000 and 3000 dimension case, almost 3 times faster. Table \ref{Tab:Acc:3000:Gene:Data} summarize the accuracy for $p=3000$ dimension case in three aspects: objective value, rank of $L$, sparsity of $S$ (Number of non-zero off-diagonal elements) for four fixed pair of $(\lambda_1,\lambda_2)$. Similar to artificial data,  SBLVGG  and  LogdetPPA generate identical results in terms of objective value and number of hidden units. However, logdetPPA suffers from the floating point problem of not being able to generate exact sparse matrix. On the other hand, SBLVGG is doing much better in this aspect.

\begin{table}[!h]
\caption{Comparison of generalization ability on gene expression data at dimension of 1000 using latent variable Gaussian graphical model (LVGG) and sparse Gaussian graphical model (SGG)}
\label{Gene:Prediction}
	\begin{center}
	\begin{tabular}{c|ccc|cc}
	\multirow{2}{*}{Exp. Number} & \multicolumn{3}{c}{LVGG} & \multicolumn{2}{c}{SGG} \\ \cline{2-6}
						      & Rank of $L$ & Sparsity of $S$ & $NLoglike$ & Sparsity of $K$ & $NLoglike$ \\ \hline
	1 & 48 & 30 & -2191.3 & 24734 & -1728.8\\
	2 & 47 & 64 & -2322.7 & 28438 & -1994.1\\
	3 & 50 & 58 & -2669.9 & 35198 & -2526.3\\
	4 & 52 & 64 & -2534.6 & 30768 & -2282.5\\
	5 & 48 & 0 & -2924.0 & 29880 & -2841.4\\
	6 & 51 & 52 & -2707.1 & 28754 & -2642.6\\
	7 & 45 & 0 & -2873.3 & 30374 & -2801.4\\
	8 & 49 & 0 & -2765.5 & 31884 & -2536.7\\
	9 & 48 & 54 & -2352.0 & 29752 & -2087.2\\
	10& 47 & 0 & -2922.9 & 29760 & -2843.5\\\hline
	\end{tabular}
	\end{center}
\end{table}

We also investigated generalization ability of latent variable Gaussian graphical selection model \eqref{Problem formulation} versus sparse Gaussian graphical model \eqref{min l1 penalized likelihood} using this data set.  A subset of the data, 1000 genes with highest variances, are used for this experiment. The 300 samples are randomly divided into 200 for training and 100 for testing.  Denote the negative log likelihood (up to a constant difference) $$NLoglike=-\log\det A + tr(A\Sigma^n),$$ where $\Sigma^n$ is the empirical covariance matrix using observed sample data and $A$ is the estimated covariance matrix based on model \eqref{Problem formulation} or model \eqref{min l1 penalized likelihood}.  It easy to see that $NLoglike$ is equivalent to negative Log-likelihood function up to some scaling. Therefore, we use $NLoglike$ as a criteria for cross-validation or prediction. Regularization parameters $\lambda_1, \lambda_2$ for model \eqref{Problem formulation} and $\lambda$ for model \eqref{min l1 penalized likelihood} are selected by 10-fold cross validation on training set. Table \ref{Gene:Prediction} shows that  latent variable Gaussian graphical selection model \eqref{Problem formulation} consistently outperform sparse Gaussian graphical model \eqref{min l1 penalized likelihood} in terms of generalization ability using criteria $NLoglike$. We also note that latent variable Gaussian graphical selection model \eqref{Problem formulation} tend to use moderate number of hidden units, and very sparse conditional correlation to explain the data. For $p=1000$, it tend to predict about $50$ hidden units, and the number of direct interconnections between observed variables are tens, and sometimes even 0. This suggests that most of the correlations between genes observed in the mRNA measurement can be explained by only a small number of latent factors.  Currently we only tested the generalization ability of latent variable Gaussian graphical selection model  using $NLoglike$. The initial result with gene expression data is encouraging. Further work (model selection and validation) will be done by incorporating other prior information or by comparing with some known gene interactions.

\section{Discussion}
Graphical model selection in high-dimension arises in a wide range of applications. It is common that in many of these applications, only a subset of the variables are directly observable. Under this scenario, the marginal concentration matrix of the observed variables is generally not sparse due to the marginalization of latent variables. A computational attractive approach is to decompose the marginal concentration matrix into a sparse matrix and a low-rank matrix, which reveals the conditional graphical model structure in the observed variables as well as the number of and effect due to the hidden variables.  Solving the regularized maximum likelihood problem is however nontrivial for large-scale problems, because of the complexity of the log-likelihood term, the trace norm penalty and $\ell_1$ norm penalty. In this work, we propose a new approach based on the split Bregman method (SBLVGG) to solve it. We show that our algorithm is at least three times faster than the state-of-art solver for large-scale problems.

We applied the method to analyze the expression of genes in yeast in a dataset consisting of thousands of genes measured over 300 different experimental conditions. It is interesting to note that the model considering the latent variables consistently outperforms the one without considering latent variables in term of testing likelihood. We also note that most of the correlations observed between mRNAs can be explained by only a few dozen latent variables. The observation is consistent with the module network idea proposed in the genomics community. It also might suggest that the postranscriptional regulation might play more prominent role than previously appreciated.

\bibliographystyle{unsrtnat}
\bibliography{ygb}
\end{document}